\documentclass[review]{elsarticle}

\usepackage{lineno}
\usepackage{amssymb}
\usepackage{latexsym}
\usepackage{color}
\usepackage{amsmath}
\usepackage{hyperref}
\usepackage{cleveref}
\usepackage{siunitx}
\usepackage{multirow}
\usepackage{subcaption}
\usepackage{amssymb}
\usepackage{tabularx}
\modulolinenumbers[5]

\journal{Journal of \LaTeX\ Templates}

\begin{document}

\begin{frontmatter}

\title{A Deep Convolutional Neural Network for the Detection of Polyps in Colonoscopy Images.}
\tnotetext[mytitlenote]{Fully documented templates are available in the elsarticle package on \href{http://www.ctan.org/tex-archive/macros/latex/contrib/elsarticle}{CTAN}.}

\author{Tariq Rahim, Syed Ali Hassan, Soo Young Shin*}
\address{Kumoh National Institute of Technology, Gumi, Gyeongbuk 39177, Republic of Korea}

\author{tariqrahim@ieee.org, Syedali@kumoh.ac.kr, wdragon@kumoh.ac.kr}

\cortext[mycorrespondingauthor]{Soo Young Shin*}
\ead{wdragon@kumoh.ac.kr}

\begin{abstract}
Computerized detection of colonic polyps remains an unsolved issue because of the wide variation in the appearance, texture, color, size, and presence of the multiple polyp-like imitators during colonoscopy. In this paper, we propose a deep convolutional neural network based model for the computerized detection of polyps within colonoscopy images. The proposed model comprises 16 convolutional layers with 2 fully connected layers, and a Softmax layer, where we implement a unique approach using different convolutional kernels within the same hidden layer for deeper feature extraction. We applied two different activation functions, MISH and rectified linear unit activation functions for deeper propagation of information and self regularized smooth non-monotonicity. Furthermore, we used a generalized intersection of union, thus overcoming issues such as scale invariance, rotation, and shape. Data augmentation techniques such as photometric and geometric distortions are adapted to overcome the obstacles faced in polyp detection. Detailed benchmarked results are provided, showing better performance in terms of precision, sensitivity, F1- score, F2- score, and dice-coefficient, thus proving the efficacy of the proposed model.
\end{abstract}


\end{frontmatter}


\section{Introduction}
Diagnosing of distinct diseases within the small intestine is a time-consuming and hectic process for physicians. This has led to the introduction of technologies such as colonoscopy and wireless capsule endoscopy \cite{rahim2019survey}. Colorectal cancer (CRC) is the second-highest cause of death by cancer worldwide with 880,792 deaths and a mortality rate of 47.60\% in 2018 reported by American Cancer Society \cite{segal2018cancer}  95\% of CRC cases start with the appearance of a growth on the inner lining of the rectum or colon, called a polyp. Various types of polyps exist including, adenoma polyps, which can worsen into CRC. CRC is curable in 90\% of cased assuming early detection \cite{chuquimia2019polyp}. Colonoscopy has emerged as minimally invasive and additional tool for investigating polyps by examining the gastrointestinal tract \cite{chuquimia2019polyp}. Colonoscopy relies on highly skillful endoscopists, and recent clinical investigations have shown colonoscopy misses 22\%– 28\% of polyps. This false negatives can lead to late diagnosis of colon cancer, resulting in a survival rate as low 10\% \cite{leufkens2012factors}.
\par  
Deep learning (DL) is a subtype of machine learning concerned with the structure and function of brain-like systems known as artificial neural networks \cite{hassan2019real}. DL plays an important role in many areas, including text recognition tasks, self-driving cars, image recognition, and healthcare. Computer vision and machine learning-based methods have revolved over several decades to automatically detect polyps \cite{bernal2017comparative, zhang2018polyp, tajbakhsh2018system}. Such systems have generally examined, hand-crafted features, such as texture, histograms of oriented gradients , color wavelets, Haar, and local binary patterns \cite{bernal2012towards, park2012colon}. More advanced algorithms have been suggested to evaluate poyp appearance based on factors such as context information \cite{tajbakhsh2015automated} and edge shape \cite{bernal2015wm}. However, the decrease in detection performance is mainly due to the similar appearance of polyp-like and polyp structures.
\par 
Convolutional neural networks (CNNs) present promising outcomes in polyp detection and segmentation. CNN features outperformed hand-crafted features in the MICCAI 2015 polyp detection challenge \cite{bernal2017comparative}. The Region-based CNN approaches, such as \textit{R-CNN} \cite{girshick2014rich}, \textit{Fast R-CNN} \cite{girshick2015fast}, and \textit{Faster R-CNN} \cite{ren2015faster} have shown promising results for object detection in natural images. Work has also been done on regression-based object detection models such as You Only Look Once (YOLO) \cite{redmon2016you} and single shot multi‐box detector (SSD) \cite{liu2016ssd}. However, recent investigations have shown that deep neural networks (DNNs), including CNNs, are extremely vulnerable to noise and perturbations \cite{qadir2019improving}. Even one single-pixel addition increases the miss detection vulnerability of current DNNs including CNNs \cite{su2019one}. Even though computer-aided detection techniques can effectively classify frames from a colonoscopy, detection of polyps remains challenging  due to significant size, appearance, and intensity variations between frames. This is a serious issue, because polyps and polyp-like objects have similar appearances in consecutive frames, leading to miss-detection even when implementing powerful models such as CNNs. Furthermore, the performance of DL approaches is highly correlated with the amount of data available for training. The lack of availability of labeled polyp images for training makes the detection and segmentation of the polyp a difficult task \cite{chao2019application}.    
\par 
This work presents a new-CNN based detection model of polyp in colonoscopy images. The proposed CNN model employs fewer hidden layers, making the model lighter and less time-consuming during training. The proposed CNN model employs MISH as an activation function in some of the hidden layers for better deep propagation of information within the CNN \cite{misra2019mish}. Data augmentation such as photometric and geometric distortions is performed due to the scarcity of annotated polyp images generated from the colonoscopy process. The rest of the paper is categorized as follows: Section 2 presents recent related work done on polyp detection in colonoscopy images using DL. In Section 3, the proposed CNN model for polyp detection in colonoscopy images is explained in detail. in Section 4, the experimental results are described in detail, along with the dataset acquisition and augmentation process. Finally, in Section 5 the paper is reviewed and concluded and future work is presented.         
\section{Related Work} 
As CNN techniques for single and multiple object detection advance, they increasingly outperform previous conventional image processing techniques \cite{zhang2019rgb}. For multiple object detection, a region-based CNN combined with a deformable part-based model has been proposed to handle feature extraction and occlusion \cite{li2018multiple}. Recently, with the progress of DL in multiple image processing applications, a CNN- based method has been introduced for polyp detection \cite{park2016colonoscopic, park2015polyp}. Due to this and related progress, CNN features outperformed hand-crafted features in the MICCAI 2015 polyp detection challenge \cite{bernal2017comparative}. A regression-based CNN model using ResYOLO combined with efficient convolution operators has been shown to successfully track and detect polyps in colonoscopy videos \cite{zhang2018polyp}. To avoid miss-detection of polyp between neighboring frames, a two-stage detector including a CNN-based object detector and a false-positive reduction unit can be applied \cite{qadir2019improving}. Automatic detection of hyperplastic and adenomatous colorectal polyps in colonoscopy images has been performed using sequentially connected encoder-decoder based CNN  \cite{bravo2020automatic}. Furthermore, automatic polyp detection in colonoscopy videos can be conducted via ensemble CNN, which learns a variety of polyp features such as texture, color, shape, and temporal information \cite{tajbakhsh2015automatic}. 
\par 
To overcome the lack of sufficient training samples for the use of pre-trained CNN on large-scale natural images, transfer learning systems have been proposed. This has been successfully implemented in various medical applications, such as automatic interleaving between radiology reports and diagnostic CT \cite{shin2015interleaved}, MRI imaging, and ultrasound imaging \cite{chen2015standard}. Furthermore, the performance of various CNN architectures based on transfer learning, such as AlexNet and GoogLeNet has been evaluated for classification of interstitial lung disease and detection of thoracic-abnormal lymph nodes \cite{shin2016deep}. Similarly, a transfer learning-based method using the deep-CNN model Inception Resnet has been used to detect polyps in colonoscopy images \cite{shin2018automatic}. Questions of whether a CNN with adequate fine-tuning can overcome the full training of the model from scratch have been answered in detail by examination of four different medical imaging applications in three  different  specialties: gastroenterology, radiology, and cardiology for the purpose of classification, detection, and segmentation \cite{tajbakhsh2016convolutional}.   
\par 
CNN has been used for decades in the field of computer vision for various applications. However, training a deep CNN model from scratch a complicated task \cite{tajbakhsh2016convolutional}. Deep CNN models require a large amount of labeled training data. This becomes a difficult requirement when large-scale annotated medical data set are unavailable. Training the models is tedious and computationally time-consuming, and becomes even more so when facing complications such as over-fitting and convergence. To overcome these issues, this work presents a form of CNN-based detection of polyps in colonoscopy images. The proposed model employs fewer hidden layers, making the model lighter and less time-consuming during training. The model uses MISH as an activation function in some of the hidden layers for better deep propagation of information within the CNN \cite{misra2019mish}. Data augmentation methods such as photometric and geometric distortions are used due to the scarcity of annotated polyp images generated from the colonoscopy process.   
\section{Proposed Deep Convolutional Neural Network (CNN) Architecture} 
Initially, the input image is divided into a grid during the training phase. Then the image is labeled using the RectLabel tool, generating a bounding box \textit{``B"} consisting of five features. The horizontal and vertical components are labeled  \textit{``x”}, and \textit{``y”}, respectively. Height and width are labeled  \textit{``h"} and \textit{``w”}, respectively. Finally, a confidence score $``{C_s}"$ is defined  for each defined grid cell. The objective function of bounding box \textit{``B"} is a bag of freebies using mean square error (MSE) to perform regression on the center coordinate points, height, and width of the box \textit{``B"}. The intersection over union (IoU) is a vital indicator for estimating the distance between the predicted truth and the ground truth \textit{``B"} and is given as generalized form as   

\begin{figure*}[t]
	\centering
	\includegraphics[width=5.0in,height=20.00in,keepaspectratio]{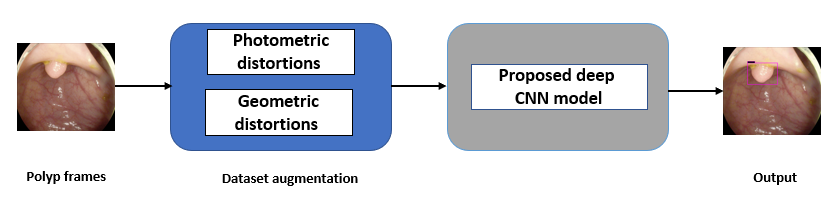}
	\caption {Flow of the proposed deep CNN model for polyp detection.}
	\label{Fig4}
\end{figure*}

\begin{figure*}[t]
	\centering
	\includegraphics[width=5.00in,height=24in,keepaspectratio]{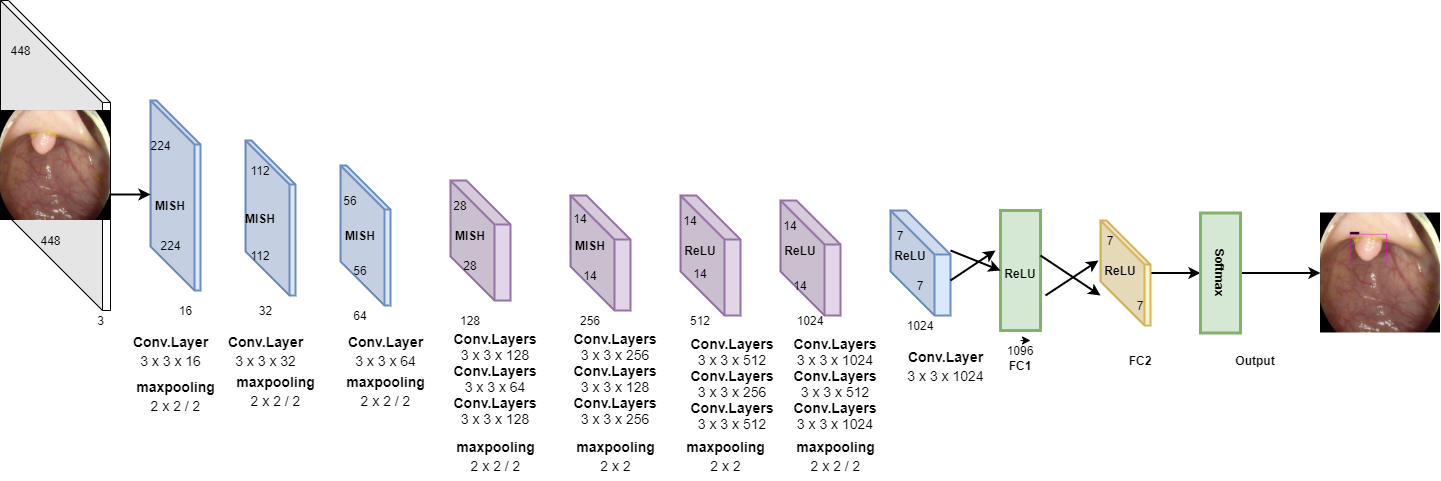}
	\caption {Architecture of the proposed deep CNN for polyp detection.}
	\label{Fig4}
\end{figure*}
\par 
\begin{equation}
I o U=\frac{|E \cap F|}{|E \cup F|}=\frac{|I|}{|U|}
\end{equation}
where \textit{``E"} and \textit{``F"} represent the predicted truth and ground truth, respectively. Here, the IoU distance $\mathcal{L}_{I o U}=1-I o U$ fulfills all properties of a metric, including the identity of indiscernibles, non-negativity, triangle inequality, and symmetry, but has a scale-invariant issue.  The cost or loss function for object detection use \textit{${l_1}$} norm and \textit{${l_2}$} norm for ${x,y,w,h}$, but due to the scale-invariant property of IoU, there is an increase in loss with respect to scaling. In the proposed deep CNN approach to polyp detection, we have implemented generalized (GIoU) \cite{rezatofighi2019generalized} as a new loss to optimize the non-overlapping \textit{``B"} in consideration of the shape and orientation of the object in \textit{``B"}. The GIoU finds the smallest convex shape $C \subseteq \mathbb{S} \in \mathbb{R}^{n}$ for two arbitrary convex shapes $E, F \subseteq \mathbb{S} \in \mathbb{R}^{n}$ followed by the calculation of the ratio between the area occupied by ${C}$ minus ${``E"}$ and ${``F"}$, divided by the total area occupied by ${``C"}$. Details for the algorithm and formulation can be found in \cite{rezatofighi2019generalized}, where a GIoU is expressed in simple mathematical form as, $G I o U=I o U-\frac{|C \backslash(E \cup F)|}{|C|}$. Furthermore, we have applied a non-maximum suppression algorithm involving $``{C_s}"$ to avoid multiple and overlapping GIoUs.
\begin{figure}[]
	\centering
	\includegraphics[width=8.80in,height=2.15in,keepaspectratio]{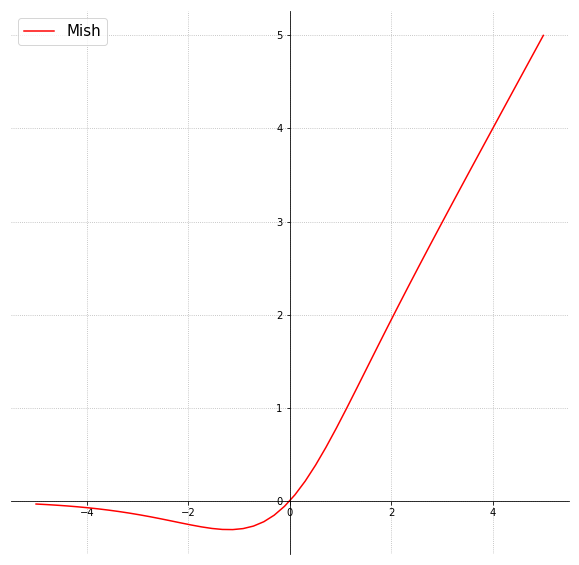}
	\caption {Mish activation function}
	\label{Fig4}
\end{figure}
\par 
Fig. 1 shows the general flow of the proposed approach for polyp detection in colonoscopy images. As shown in Fig. 2, the proposed deep CNN consists of 16 convolutional layers, two fully connected layers, and a softmax layer. To lessen computational complexity and improve hierarchical image features, \textit{maxpooling} is used for the first 15 convolutional layers. For better image feature extraction, different sizes of \textit{convolution kernels} are employed, with a stride of 2. In the proposed model, we have implemented \textit{Mish} \cite{misra2019mish}, which is a self-regularized smooth non-monotonic activation function, in the first 15 convolutional layers. This implementation was done after extensive  trials to find the best matching position of the activation function. As observed in Fig. 3, \textit{Mish} is an unbounded above result in avoiding saturation due to capping. This may normally lead to slow training, i.e., near-zero gradients. A better gradient flow and smooth propagation of information across deeper layers are achieved by the infinite order of continuity and a small allowance of negative values, in comparison to a strictly bounded rectified linear unit (ReLU) as an activation function. MISH can be expressed mathematically as:  
\begin{equation}
f(x)=x \cdot \tanh (\zeta(x))
\end{equation}
where, $\zeta(x)=\ln \left(1+e^{x}\right)$ is softplus activation \cite{misra2019mish}.    
\par 
In the last layers, ReLU is used as an activation function to reduce the likelihood of gradient vanishing and achieve the sparsity. Flattening is done by two fully connected layers to yield a single continuous linear vector followed by \textit{softmax} or the regression layer to generate the required output. The approach of using \textit{Mish} and ReLU as an activation function results in smooth propagation of information across deeper layers. MISH helps to avoid capping, and ReLU prevents the gradient from vanishing.  
\par 
The proposed deep CNN is trained with a similar concept of multi-layer perceptions i.e., a back-propagation algorithm which minimizes the cost function concerning the unknown weights ${``W"}$:   
\begin{equation}
\mathcal{L}=-\frac{1}{|L|} \sum_{i}^{|L|} \ln \left(p\left(m^{i} | L^{i}\right)\right)
\end{equation}
where $|L|$ represents the number of training images, $p\left(m^{i} | L^{i}\right)$ represents the probability that $L^{i}$ is accurately classified, and $L^{i}$ represents the $i^{t h}$ training image with the associated label $m^{i}$. We have applied stochastic gradient descent (SGD) as an optimizer, which minimizes the cost function over the whole training data set along with the cost over mini-batches of data. If $W_{j}^{t}$ represents the weights in the $j^{t h}$ convolutional layer at ${t}$ iteration, and $\hat{\mathcal{L}}$ represents the cost over a mini-batch of size ${M}$, then in the next iteration the updated weights are calculated as given below:  
\begin{equation}\begin{aligned}
\gamma^{t} &=\gamma^{\lfloor t M /|X|\rfloor} \\
V_{j}^{t+1} &=\mu V_{j}^{t}-\gamma^{t} \eta_{j} \frac{\partial \hat{\mathcal{L}}}{\partial W_{j}} \\
W_{j}^{t+1} &=W_{j}^{t}+V_{j}^{t+1}
\end{aligned}\end{equation}
where $\eta_{j}$ is the learning rate of the $j^{t h}$, $\mu$ is the momentum indicating the previously updated weight contribution in the current iteration, and $\gamma$ represent the scheduling rate which after each epoch decreases the learning rate ${\eta}$. The simulation parameters used for the proposed deep CNN are given in Table. 1.

\begin{table}[h]
	\renewcommand{\arraystretch}{0.6}
	\caption{Simulation parameters used for deep CNN model}
	\resizebox{12.00cm}{!}{
		\begin{tabular}{|c|c|}
			\hline
			\textbf{Network parameters} & \textbf{Configuration values}                                                \\ \hline
			Input image dimension       & 448 X 448                                                                    \\ \hline
			Learning rate               & 0.0001                                                                       \\ \hline
			Optimizer                   & \begin{tabular}[c]{@{}c@{}}Stochastic gradient \\ descent (SGD)\end{tabular} \\ \hline
			Momentum                    & 0.9                                                                          \\ \hline
			Bath size                   & 32                                                                           \\ \hline
			Iterations (t)              & 10,000                                                                       \\ \hline
		\end{tabular}
	}
\end{table}
\begin{figure}[h]
	\begin{subfigure}{0.33\textwidth}
		\centering
		\includegraphics[width=3.5cm, height= 2.2cm]{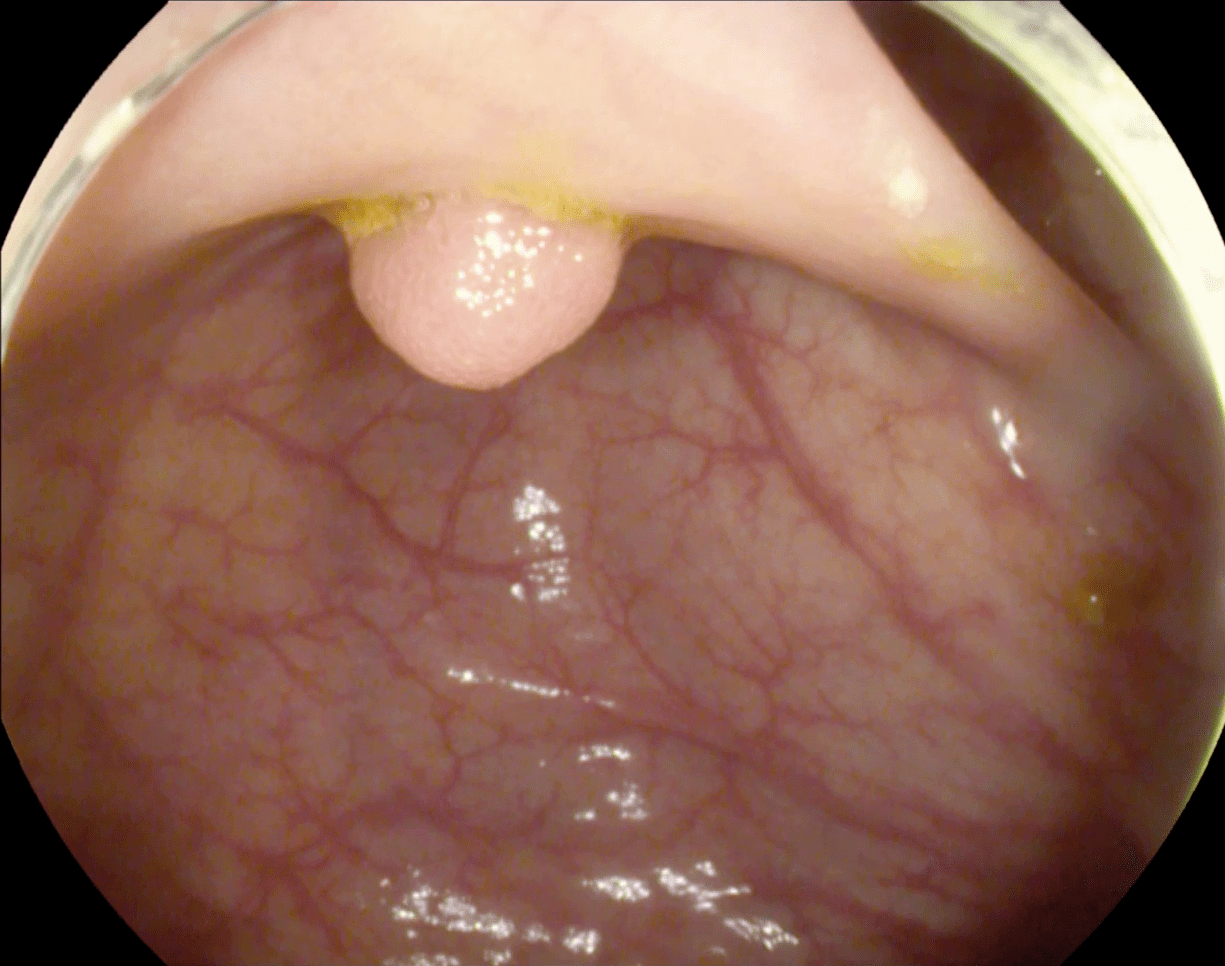}
		\caption{}
		\label{fig_med_pdr}
	\end{subfigure}%
	~\begin{subfigure}{0.33\textwidth}
		\centering
		\includegraphics[width=3.5cm, height= 2.2cm]{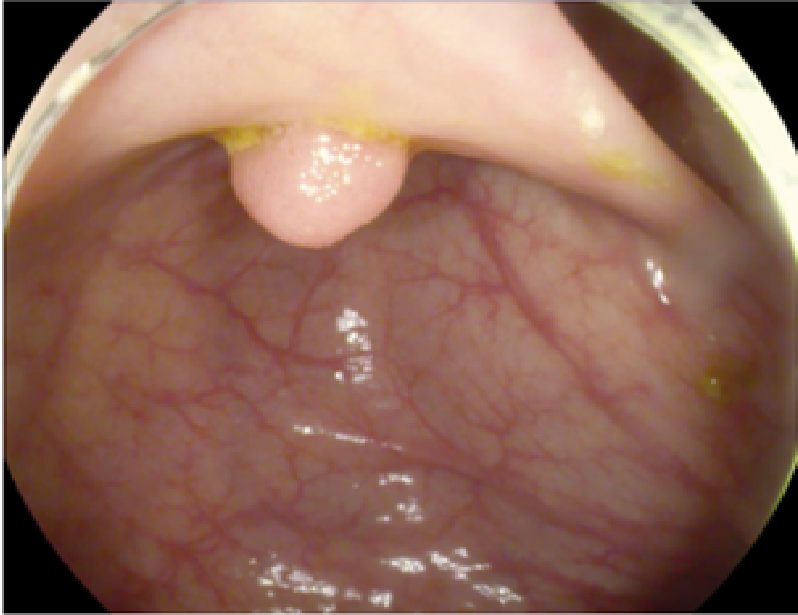}
		\caption{}
		\label{fig_med_hops}
	\end{subfigure}%
	~\begin{subfigure}{0.33\textwidth}
		\centering
		\includegraphics[width=3.5cm, height= 2.2cm]{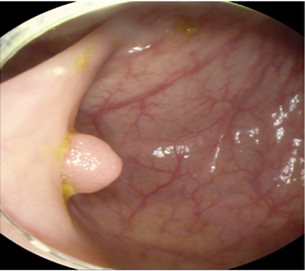}
		\caption{}
		\label{fig_med_ct}
	\end{subfigure}
	\\
	\begin{subfigure}{0.33\textwidth}
		\centering
		\includegraphics[width=3.5cm, height= 2.2cm]{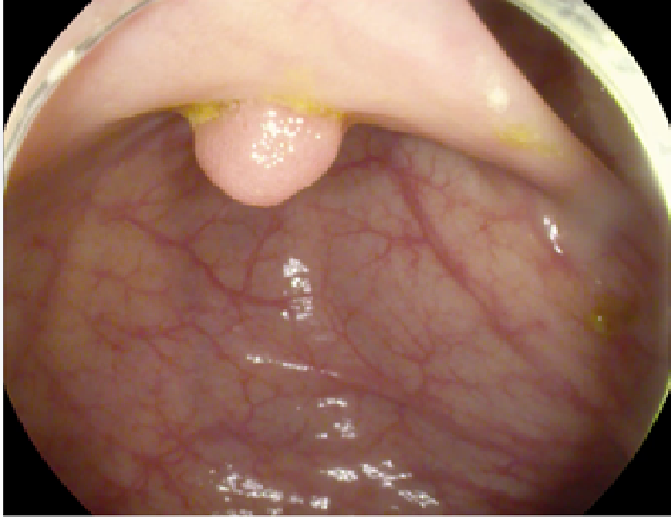}
		\caption{}
		\label{fig_med_pdr}
	\end{subfigure}%
	~\begin{subfigure}{0.33\textwidth}
		\centering
		\includegraphics[width=3.5cm, height= 2.2cm]{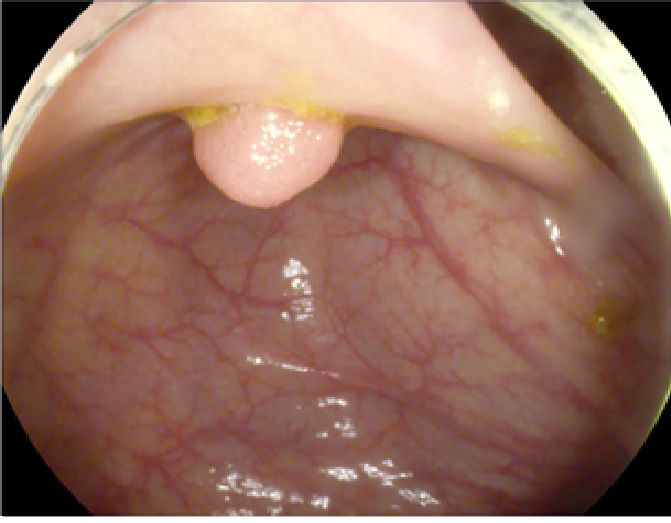}
		\caption{}
		\label{fig_med_hops}
	\end{subfigure}%
	~\begin{subfigure}{0.33\textwidth}
		\centering
		\includegraphics[width=3.5cm, height= 2.2cm]{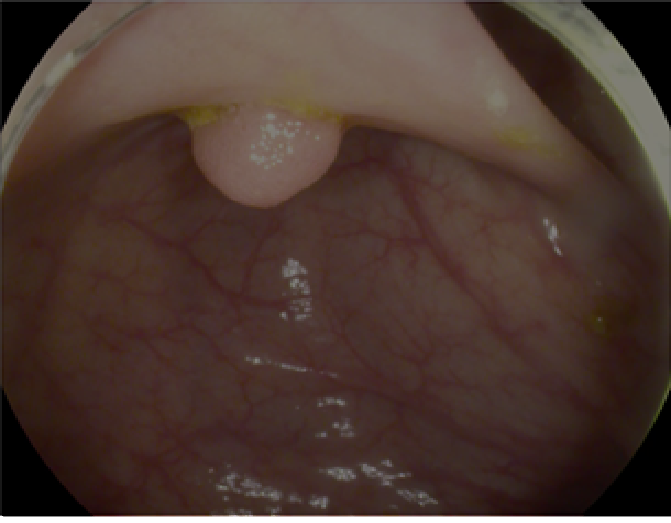}
		\caption{}
		\label{fig_med_ct}
	\end{subfigure}%
	\\
	\begin{subfigure}{0.33\textwidth}
		\centering
		\includegraphics[width=3.5cm, height= 2.2cm]{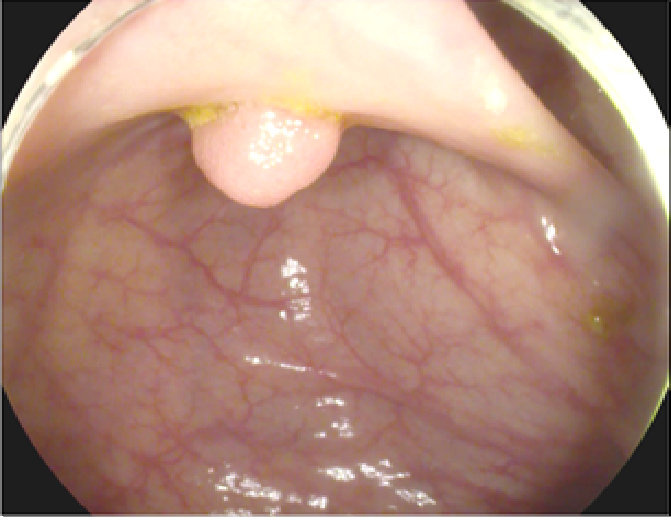}
		\caption{}
		\label{fig_med_pdr}
	\end{subfigure}%
	~\begin{subfigure}{0.33\textwidth}
		\centering
		\includegraphics[width=3.5cm, height= 2.2cm]{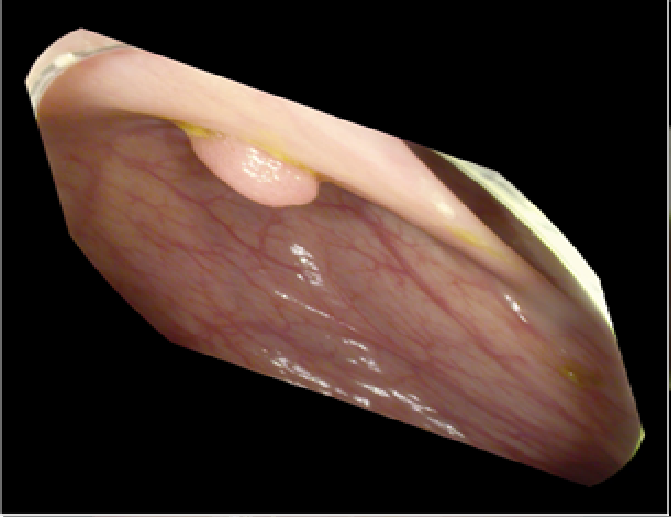}
		\caption{}
		\label{fig_med_hops}
	\end{subfigure}%
	~\begin{subfigure}{0.33\textwidth}
		\centering
		\includegraphics[width=3.5cm, height= 2.2cm]{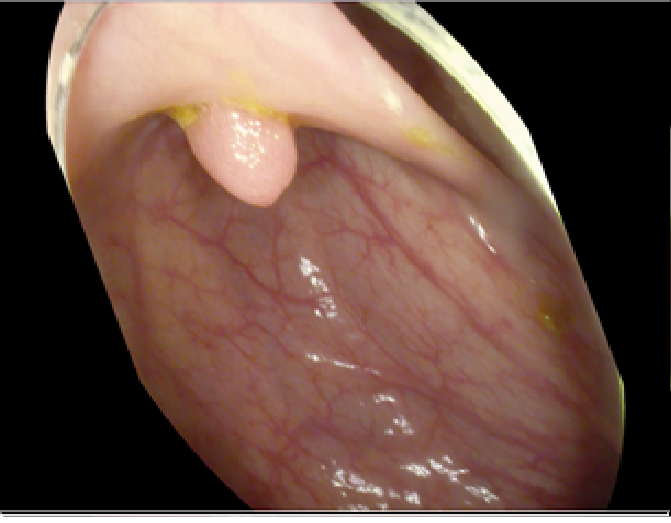}
		\caption{}
		\label{fig_med_ct}
	\end{subfigure}%
	\caption{Image from the dataset with photometric and geometric augmentation. (a) original frame of polyp, (b) noisy polyp frame with $\sigma=1.1$, (c) rotated polyp frame with \ang{90}, (d) polyp frame with 15.00\% zoom in, (e)  polyp frame with 15.00\% zoom out, (f) dark polyp frame, (g) bright polyp frame, (h) sheared polyp frame by y-axis, (i) sheared polyp frame by x-axis.}
	\label{Fig:Results_Collisions}
\end{figure}
\section{Experimental Results and Discussion}
This section detail the data set specifications and experimental results generated by implementing the proposed deep CNN for the detection of polyps in colonoscopy images. 
\subsection{Dataset Specifications and Augmentation}
The study used a publicly available dataset of polyp-frames obtained from the ETIS-Larib database \cite{silva2014toward}, containing 196 polyp images. These images were obtained from 34 different colonoscopy videos of 44 different polyps with various appearances and sizes, having a resolution of $1225\times966$ pixels. The ground truth of polyp areas for polyp datasets is determined by expert video endoscopists. A CNN model trained with such a small amount of data is likely to be meaningless and unstable, so data augmentation was performed on the polyp dataset. Data augmentation had to be performed on the colonoscopy images by considering vivid variations. Otherwise over-fitting would have occurred. In a colonoscopy imagery, polyps exhibits large variations in location, color, and scale. Moreover, variations in brightness and definition also occur due varrying the view-point of the camera. Therefore, in addition to photometric distortions and geometric distortions, we also have considered zooming, shearing, and altering brightness as strategies for data augmentation.   
\par
For photometric distortions, we controlled brightness and contrast as an enhancement, while blurring by adding noise with a standard deviation $(\sigma)$ of 1.0. Similarly, for geometric distortions, clock-wise rotation of the polyp images with angles of \ang{90}, \ang{180}, and \ang{270} were performed. Zoom-in and zoom-out with zooming parameters such as 30.00\% and 10.00\% were performed to obtain different scales of polyp images. Lastly, shearing for both the x-axis and the y-axis was performed to shear the images from left to right and top to bottom, respectively. Fig. 4 shows photometric and geometric forms of image augmentation. In this way, we augmented the data set of the ETIS-Larib database from 196 polyp images to 2,156 images, which is more suitable for training the proposed deep CNN model.    

\subsection{Performance Metrics for Evaluations}
The metrics used to evaluate the detection of polyps within colonoscopy frames in this work are the same as those used in the MICCAI 2015 challenge \cite{bernal2017comparative}. The output obtained using the proposed model has rectangular shaped coordinates (\textit {x, y, w, h}). The following parameters are defined as follows:
\par
\textbf{True Positive (TP):} True output detection if the detected centroid falls within the polyp ground truth. For multiple true output detection within the same frame and of the same polyp, TP is counted as one.  
\par
\textbf{True Negative (TN):} True detection, i.e., negative frames (frames without polyps) yielding no detection output. 
\par 
\textbf{False Positive (FP):} False detection output where the detected centroid falls outside the polyp ground truth. 
\par 
\textbf{False Negative (FN):} False detection output, i.e, polyp is missed in a frame having a polyp.
\par 
Employing the above parameters, we can compute the following performance metrics to efficiently evaluate  the performance of the proposed deep CNN model. 
\par 
\textbf{Precision:} This metric computes how precisely the model is detecting a polyp within an image 
\begin{equation}
\text {Precision }(\text {Pre})=\frac{T P}{T P+F P} \times 100
\end{equation}
\par 
\textbf{Sensitivity:} This metric is also called recall or True Positive Rate and computes the proportion of the actual polyps that were detected correctly
\begin{equation}
\text {Sensitivity }(\text {Sen})=\frac{T P}{T P+F N} \times 100
\end{equation}
\par 
\textbf{F1- score and F2- score:} F1 and F2- score is simply the harmonic mean between precision and sensitivity , in a range of [0, 1]. Both scores are recognized to balance the precision and sensitivity. The F1- score is given as:
\begin{equation}
F 1-\text {score}=\frac{2 \times \text {Sen } \times \text { Pre}}{\text {Sen }+\text {Pre}} \times 100
\end{equation}
while the F2- score can be calculated as:
\begin{equation}
F 2- score=\frac{5 \times \text {Pre} \times \text {Sen}}{4 \times \text {Pre}+\text {Sen}}
\end{equation}
\par 
\textbf{Dice Coefficient:} This metric is used for pixel-wise result comparison between ground truth and predicted detection that ranges [0, 1], and is given as:
\begin{equation}
\text { Dice coefficient }(E, F)=\frac{2 \times|E \cap F|}{|E|+|F|}=\frac{2 \times T P}{2 \times T P+F P+F N}
\end{equation}

\begin{figure*}[t]
	\centering
	\includegraphics[width=10.80in,height=3.30in,keepaspectratio]{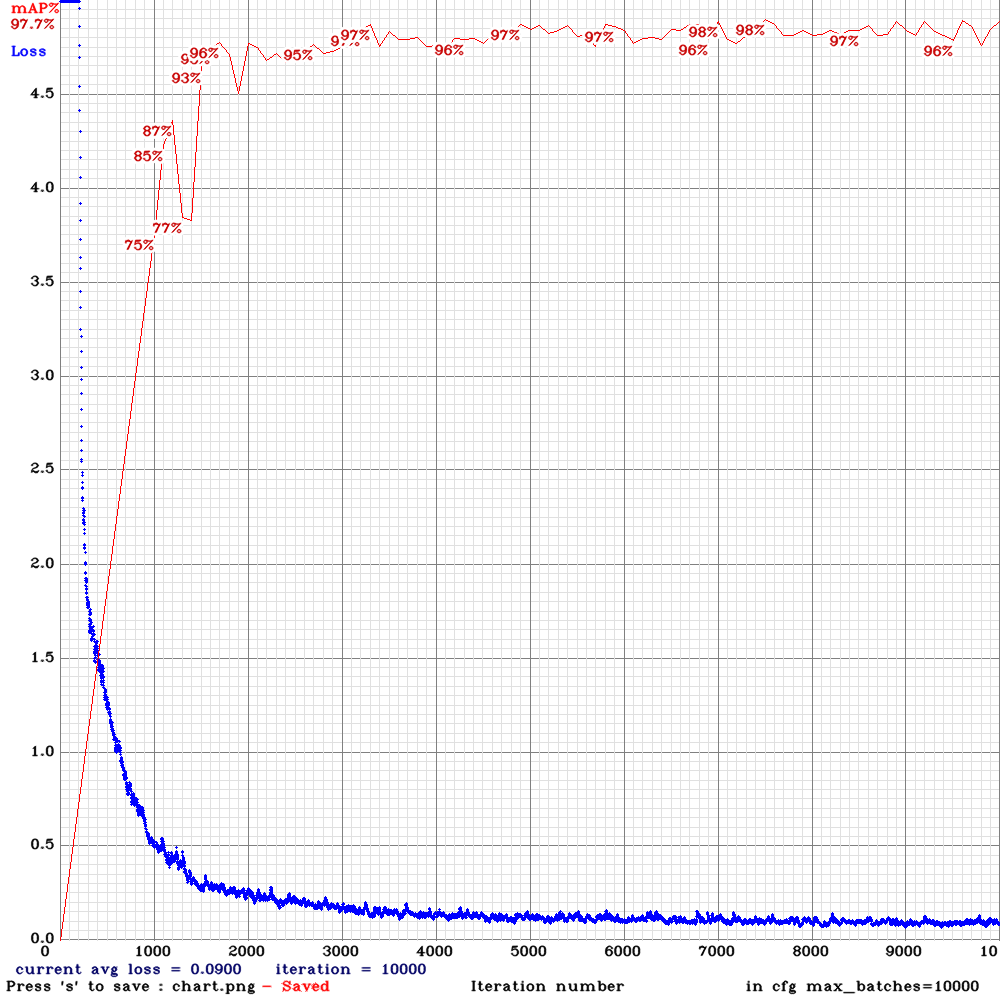}
	\caption {Training phase of the proposed deep CNN model.}
	\label{Fig4}
\end{figure*}
\subsection{Polyp Frames Evaluation}
This section reports the polyp detection performance of the proposed CNN model. For implementation of the model, 80.00\% and 20.00\% of the 2,156 augmented polyp frames were  used for training and testing, respectively. Fig. 5 shows the real-time training phase of the proposed model, where 10,000 iterations were run to achieve the best weights. The model was trained using the simulation parameters as given in Table. 1, both for the non-augmented ETIS-Larib database \cite{silva2014toward} containing 196 poly images, and the augmented data set, for fair performance comparison. A high mean average precision of 97.70\% with an MSE of 0.900 was obtained in the early iterations, resulting in the best weights for testing purposes.  
\par 
The results listed in Table. 2 using the proposed deep CNN model, show better performance, with high values for precision, the F1-score, the F2-score, and the dice coefficients. Note that the low sensitivity or recall values is an indication of better polyp detection performance in the proposed model. As observed in Table. 2, the proposed model is compared to other works \cite{shin2018automatic} that showed better performance for both the non-augmented and augmented case. For the non-augmented data set of ETIS-Larib, the generated TP, FP, and FN values were 90, 35, and 51, respectively. Similarly, 20.00\% of the augmented data set was employed for testing purposes, generating TP, FP, and FN values of 340, 20, and 70, respectively.  

\begin{table}[h]
	\renewcommand{\arraystretch}{0.8}
	\caption{Detection performance comparison of the proposed deep CNN model on ETIS-Larib database without(w/0) and with augmentation strategies.}
	\resizebox{12.00cm}{!}{
		\begin{tabular}{|c|c|c|c|c|}
			\hline
			\multirow{2}{*}{\textbf{Data set}}                                                                             & \multicolumn{4}{c|}{\textbf{Performance metrics}}                                                                                      \\ \cline{2-5} 
			& \multicolumn{2}{c|}{\textbf{\cite{shin2018automatic}} (\%)} & \multicolumn{2}{c|}{\textbf{\begin{tabular}[c]{@{}c@{}}Propose deep \\ CNN model (\%)\end{tabular}}} \\ \hline
			\multirow{5}{*}{\textbf{\begin{tabular}[c]{@{}c@{}}Non-augmented\\ ETIS LARIB \\ database (196)\end{tabular}}} & \textbf{Pre}                & 48.00  & \textbf{Pre}                                         & \textbf{72.00}                           \\ \cline{2-5} 
			& \textbf{Sen}                & 39.40  & \textbf{Sen}                                         & \textbf{63.82}                           \\ \cline{2-5} 
			& \textbf{F1- score}          & 43.30  & \textbf{F1- score}                                   & \textbf{67.66}                           \\ \cline{2-5} 
			& \textbf{F2- score}          & 40.90  & \textbf{F2- score}                                   & \textbf{65.30}                           \\ \cline{2-5} 
			& \textbf{Dice-coefficient}   & NA     & \textbf{Dice-coefficient}                            & \textbf{0.676}                           \\ \hline
			\multirow{5}{*}{\textbf{\begin{tabular}[c]{@{}c@{}}Augmented ETIS\\ LARIB database\\ (2,156)\end{tabular}}}    & \textbf{Pre}                & 91.40  & \textbf{Pre}                                         & \textbf{94.44}                           \\ \cline{2-5} 
			& \textbf{Sen}                & 71.20  & \textbf{Sen}                                         & \textbf{82.92}                           \\ \cline{2-5} 
			& \textbf{F1- score}          & 80.00  & \textbf{F1- score}                                   & \textbf{88.30}                           \\ \cline{2-5} 
			& \textbf{F2- score}          & 74.50  & \textbf{F2- score}                                   & \textbf{85.00}                           \\ \cline{2-5} 
			& \textbf{Dice-coefficient}   & NA     & \textbf{Dice-coefficient}                            & \textbf{0.88}                            \\ \hline
		\end{tabular}
	}
\end{table}

\begin{figure*}[t]
	\centering
	\includegraphics[width=9.0in,height=2.4in,keepaspectratio]{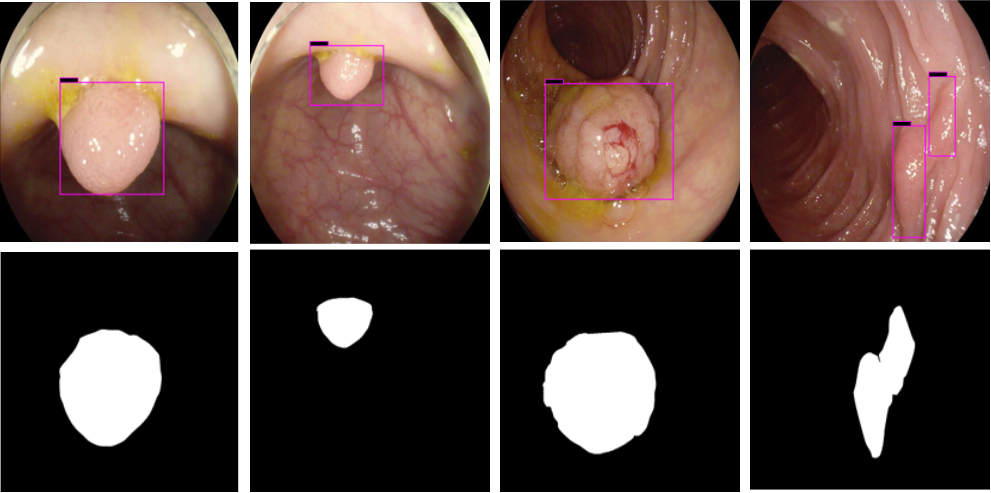}
	\caption {Example of accurate detection along with the correct ground truth using deep CNN model. The first row shows the detection results  for different polyp from data augmentation process. The second shows the ground truth images of test images.}
	\label{Fig4}
\end{figure*}
\par 
The results shown in Fig. 6 are generated using the proposed deep CNN model on the augmented data set. It can be observed that the proposed  model shows better polyp detection performance. As illustrated in Fig. 6, polyps within a frame can be identified at multiple positions, and as noted above in this case, the TP for detection is considered to be 1. The proposed deep CNN model performed better than other benchmark results in terms of the performance metrics listed above, as shown in Table. 2 and Fig. 6. 

\begin{figure*}[t]
	\centering
	\includegraphics[width=18.30in,height=4.0in,keepaspectratio]{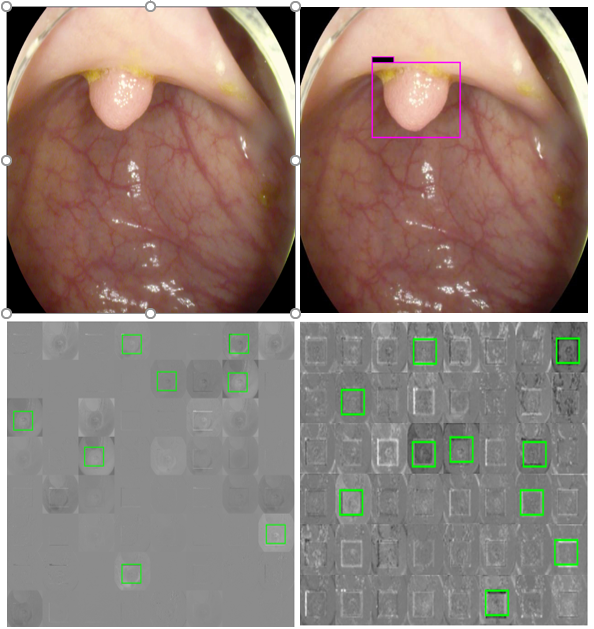}
	\caption {Test polyp channel activation visualization of CNN after training of the proposed deep CNN model. Top left: test polyp images, Top right: detection output of the test image, Bottom left: activation channel for single layer, Bottom right: activation channel for deep layer (both shown by  green rectangular box).}
	\label{Fig4}
\end{figure*}     
\par 
For single and deep layer of the proposed model, we have shown channel activation representing the convolutional kernels accurately detected the polyp. Fig. 7 shows different bright and dark parts corresponding to the spatial property of the object within the test images for single and deep layers. The \textit{top left} is the test polyp image followed by \textit{top right} detection output generated by proposed deep CNN model. The \textit{bottom left} shows the single layer activation channel whole \textit{bottom right} shows the deep layer for deeper feature analysis represented by green rectangular boxes. It can be observed in Fig. 7, that both single and deep layers are extracting polyp features with a high score, resulting in high polyp detection.  
\begin{table}[]
	\renewcommand{\arraystretch}{0.90}
	\caption{Performance comparison of the proposed deep CNN model on ETIS-Larib database with other methods.}
	\resizebox{12.00cm}{!}{
		\begin{tabular}{|c|c|c|c|c|c|}
			\hline
			\multirow{2}{*}{\textbf{\begin{tabular}[c]{@{}c@{}}Implemented \\ methods\end{tabular}}} & \multicolumn{5}{c|}{\textbf{Performance metrics}}                                                                     \\ \cline{2-6} 
			& \textbf{Pre (\%)} & \textbf{Sen (\%)} & \textbf{F1- score (\%)} & \textbf{F2- score (\%)} & \textbf{Dice-coefficient} \\ \hline
			\textbf{UNS-ULCAN}                                                                       & 32.70             & 52.80             & 40.40                   & 47.10                   & NA                        \\ \hline
			\textbf{OUS}                                                                             & 69.70             & 63.00             & 66.10                   & 64.20                   & NA                        \\ \hline
			\textbf{CUMED}                                                                           & 72.30             & 69.20             & 70.70                   & 69.80                   & NA                        \\ \hline
			\textbf{\begin{tabular}[c]{@{}c@{}}Proposed deep\\ CNN model\end{tabular}}               & \textbf{94.44}    & \textbf{82.92}    & \textbf{88.30}          & \textbf{85.00}          & \textbf{0.88}             \\ \hline
		\end{tabular}
	}
\end{table}
\subsection{Benchmark Performance with Other Approaches}
The performance of the proposed deep CNN model was benchmarked against the 2015 MICCAI challenge \cite{bernal2017comparative}, as the dataset used was the same. The top three experimental results from each team in the challenge UNS-UCLAN, OUS, and CUMED were selected for benchmarking. These results were selected because CNN has been used for learning end-to-end detection of the polyp. The UNS-UCLAN team \cite{bernal2017comparative} used three CNNs for the extraction of features on multiple spatial scales, followed by a classification approach with a multi-layer perception network. AlexNet, a CNN-based model was adopted with a conventional sliding window to perform patch-based classification \cite{shin2018automatic}. The CUMED team used a segmentation approach based on CNN \cite{silva2014toward}, where classification was conducted pixel-wise along with a ground truth mask.
\par 
As shown in Table. 3, the generated results from the proposed model using the augmented dataset outperform the other team's methods on several metrics, including precision, sensitivity, F1- score, F2- score, and dice-coefficient. As DL-based methods employ different computers with different specifications, it is hard to benchmark detection processing directly. In our work, the dataset was trained and tested on NVIDIA Titan RTX GPUs to reduce processing time. Compared to the other studies listed in Table. 3 the mean detection processing time  0.6 sec per frame. This is slightly greater than that in competing models, but the increased processing time comes with better performance.   

\section*{Conclusion}
In this paper, we presented a computerized DL-based detection model for colonic polyps. A deep CNN model consisting of 16 convolutional layers with two full-connected layers, and a Softmax layer, was implemented with different kernel sizes in the same hidden layer being employed. Moreover, two different activation functions MISH and ReLU were implemented for the first time to provide deeper propagation of information, better self-regularization, and better capping avoidance. The scale invariance issue related to IoU was addressed by adopting a GIoU that is robust under rotation and shape variation. Furthermore, photometric and geometric strategies were used for data augmentation, thus overcoming the image scarcity issue. We provided a detailed benchmark performance comparison of our detection output, which outperforms the other approaches in performance metrics such as precision, sensitivity, F1- score, F2- score, and dice-coefficient.  
\section*{Acknowledgments}
This work was supported by Priority Research Centers Program through the National Research Foundation of Korea (NRF) funded by the Ministry of Education, Science and Technology”(2018R1A6A1A03024003). 


\bibliography{mybibfile}

\end{document}